# Data Analytics using Ontologies of Management Theories: Towards Implementing "From Theory to Practice"

Henry M. Kim[1], Jackie Ho Nam Cheung[2], Marek Laskowski[1], Iryna Gel[3]

[1]*York University, Toronto, Ontario Canada*
hmkim@yorku.ca
marlas@yorku.ca

[2]*Kingsdale Shareholder Services, Toronto, Ontario Canada*
jcheung@kingsdaleshareholder.com

[3]*Canadian Tire, Toronto, Ontario Canada*
iryna.gel@cantire.com

Abstract

*We explore how computational ontologies can be impactful vis-à-vis the developing discipline of "data science." We posit an approach wherein management theories are represented as formal axioms, and then applied to draw inferences about data that reside in corporate databases. That is, management theories would be implemented as rules within a data analytics engine. We demonstrate a case study development of such an ontology by formally representing an accounting theory in First-Order Logic. Though quite preliminary, the idea that an information technology, namely ontologies, can potentially actualize the academic cliché, "From Theory to Practice," and be applicable to the burgeoning domain of data analytics is novel and exciting.*

## 1. Introduction

There has been tremendous interest in "data science," "data analytics," "Big Data," and "data-driven decision-making" [1][2][3][4]. Extending these ideas to the scientific realm, Microsoft researchers are calling the capability to analyze large volumes of data as the basis of a fourth basic research paradigm for understanding nature, one that proceeds experimental, theoretical, and simulation-based paradigms [5].

Sequencing the human genome was an early, high-profile scientific application of this capability [6]. A key data analytics techniques used in the field of genomics and other medical sciences is use of ontologies [7, pg. 1251]:

> *In the search for what is biologically and clinically significant in the swarms of data being generated by today's high-throughput technologies, a common strategy involves the creation and analysis of 'annotations' linking primary data to expressions in controlled, structured vocabularies, thereby making the data available to search and to algorithmic processing. The most successful such endeavor measured both by numbers of users and by reach across species and granularities, is the Gene Ontology (GO).*

In addition, the National Center of Biological Ontology [8] received $18M in National Institutes of Health (NIH) funding in the US. In medical sciences, use of ontologies is an acknowledged data analytics technique. There may be a parallel opportunity in management research and practice to use ontologies to realize the benefits of data analytics. In this paper, we explore that opportunity.

First, we present background context: One, what we mean when we say "ontology," and two, why such ontologies work in the medical sciences domain and how that insight can be applied to management research and practice. Next, we present a case study development of an ontology for the accounting domain that is useful for analytics. Finally, we summarize the case study, and discuss current limitations and opportunities of our approach.

## 2. Background

According to Merriam-Webster, ontology is "a branch of metaphysics dealing with the nature and relation of being" [9]. Computer Science and Information Systems fields have co-opted that term and overloaded it with two different meanings. An ontology *of* information systems is used to analyze—employing philosophical argumentation upon natural language representations—the grammars [10] and languages (e.g. ER Models [11], UML) used to develop information systems. An ontology *for* information systems underlies software applications that use ontologies as data/knowledge models to represent and reason—employing automated or semi-automated inference upon formal or semi-formal representations—about entities and their properties, relationships, constraints and behaviors [12] within a shared domain of interest [13]. According to Kishore *et al.* [14], An ontology of information systems is a philosophical ontology; an ontology for information systems is a computational ontology (aka visual ontology [15]). The ontology-driven analytics approach described in this paper develops computational ontologies.

Computational ontologies have been used, for example, in enterprise modeling [16] and knowledge management [17]. However, in terms of research funding and scope, their use in such management domains pales in comparison to their use in medical sciences. To understand why, it is instructive to examine Merriam-Webster's second definition of ontology: "a particular theory about the nature of being or the kinds of things that have existence" [9]. Ontologies are operationalized theories; how they are operationalized determines whether they are philosophical or computational. Theories in medical sciences are within the realm of natural sciences; most management theories, within social sciences. Theories in medical sciences are more testable, formal, and generalizable—i.e. a wide audience would agree upon the same meanings for terms—than most theories in management. Given that by definition computational ontologies are formal and sharable (afforded by generalizability), it is reasonable to argue that computational ontologies of, say, saliva [18] would find more real-life utility than computational ontologies of, say, organizational structure [16] and knowledge management processes [17].

Though not as formal in representation as natural science theories, theories in social sciences are nevertheless rigorously and systematically developed. Social science fields establish standards for causal inference, explanation, prediction, and generalization of theories [19]. Yet, inasmuch as medical science theories are operationalized as (computational) ontologies[1], social science oriented management theories seldom are[2]. There is a complementary work to organize constructs that form social science theories and prepare them for public access [20], but that endeavor develops sharable, informal representations, not formal representations for automated inference. Next, we demonstrate a case study operationalization of a theory in the next section to capitalize on this research opportunity.

## 3. Case Study Operationalization of an Accounting Theory

Engineering of a (computational) ontology should be like development of a theory. What is striking, however, is how elegantly aspects of the TOVE Ontological Engineering Methodology [21] map onto aspects of management theory development.

In TOVE, the Motivating Scenario is a detailed narrative about a specific enterprise which seeks to use an ontology. A paper that theorize about individual auditor decision making [22] fits nicely as a motivating scenario for our Auditor Decision-Making Ontology.

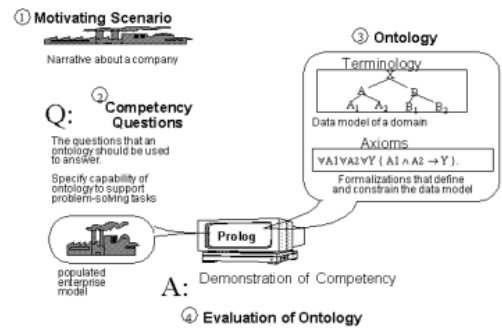

Figure 1: TOVE Ontological Engineering Methodology [21, pg. 110]

In *Joint Effects of Principles-Based versus Rules-Based Standards and Auditor Type in Constraining Financial Managers' Aggressive Reporting* [22], Jamal and Tan state hypotheses about how the auditor orientation and type of standard used in the audit affect actions of a financial manager like a CFO, who receives auditor advice prior to preparing and releasing a financial report. When the manager prefers to be opportunistic in their treatment of accounting items, the auditor may negotiate with them to heed the audit advice. When the auditor can and is willing to negotiate firmly, the manager may not prepare as opportunistic a report.

---

[1] Heretofore, the term ontology, if unqualified, refers to computational ontologies
[2] For instance, of the 98 Information Systems theories summarized at http://is.theorizeit.org/wiki/Main_Page , none have been represented as ontologies as of March 27, 2016

Then, competency questions are delineated into informal and formal competency questions. Informal Competency Questions are represented in natural language and posed using the vernacular of the motivating scenario. Generalized research questions can serve as top-level informal competency questions that spawn other questions. One example is [22, pg. 1331]:

*Can the nature of accounting standards 'undermine or support the degree to which the auditor can enforce his/her preferred stand'?*

Informal competency questions are hierarchically decomposed to more specific questions. At the bottom of this hierarchy are questions that provide an objective answer like "Is 'ifrs'[3] an accounting standard?" or "Is 'nonopportunistic' the preferred stand for auditor 'john_jones'?"

Objective and answerable informal competency questions can be posed in parallel using an ontology. For our work, they can be expressed as Formal Competency Questions in predicate form, as follows. Note that "preferred treatment" is used instead of "preferred stand."

$$holds(accounting\_standard(ifrs), \sigma) \quad (1)$$

$$holds(enforces\_preferred\_treatment(john\_jones, non\_opportunistic), \sigma). \quad (2)$$

$holds(\bullet, \sigma)$ and $occurs(\bullet, \sigma)$ are base representations of the situation calculus [23]. Any statement in $\bullet$ is valid, or "holds," in a given situational context $\sigma$. Also, a statement that signifies some action in $\bullet$ is said to "occur" in a given context $\sigma$. Situational context is an abstract concept representing the totality of relevant information needed for statements in $\bullet$ to be true. If the context materially changes, statements in $\bullet$ may not hold. Say that John Jones tended to enforce a nonopportunistic accounting treatment on his clients in his earlier years but much less so later in his career. Then,
$holds(enforces\_preferred\_treatment(john\_jones, nonopportunistic), \sigma_1)$ and
$\neg holds(enforces\_preferred\_treatment(john\_jones, nonopportunistic), \sigma_2)$ can be true as long as $\sigma_1 \neq \sigma_2$.

First-Order Logic statements can be proven true or false. An axiom is a type of statement needed to perform proofs. An axiom is comprised of predicates, and operators (e.g. ∧, ∨) and qualifiers (∀, ∃) permitted. Our ontologies are represented using First-Order Logic and are comprised of:

- Terminology: a model that lists predicates and the relationships between them
- Axioms: definitions of predicates and constraints on their use

Representations needed to prove (1) and (2) and others constitute our Auditor Decision-Making Ontology. These representations are developed in part by sharing representations from the YODA@ssb BDI (Belief-Desires-Intentions) Ontology, which is shown below.

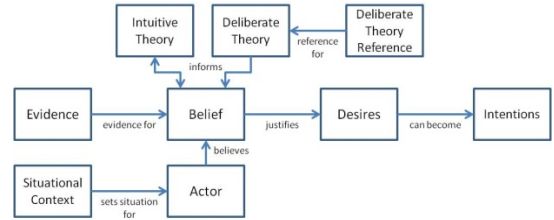

**Figure 2: YODA@ssb Belief-Desires-Intentions Ontology Data Model**

Jamal and Tan provide explicit guidance for ontology development in the form of their hypotheses. For example [22, pg. 1331-2]:

H1a: When accounting standards are rules-based, managers' likelihood of adopting an opportunistic accounting treatment will not differ as a function of whether the auditor is rules-oriented, client-oriented, or principles-oriented.
H1b: When accounting standards are principles-based, managers' likelihood of adopting an opportunistic accounting treatment will be lower when the auditor is principles-oriented than when the auditor is either rules-oriented or client-oriented.

The independent variables—type of accounting standard and orientation of auditor—characterize the auditor and their audit, while the dependent variable—the likelihood of adopting an opportunistic treatment—characterizes the manager. In Jamal and Tan's experimental design, values for independent variables are trivially provided—e.g. a respondent might be told "assume that your auditor is rules-based and you are to use rules-based GAAP"—as experimental manipulation. As for the dependent variable, respondents are told they are financial managers under a scenario about how to treat a lease and told to write a memo about their treatment. The respondents are evenly divided among the 2 (rules- and principles-based standard) X 3 (rules-, principles-, and customer-oriented auditor) manipulations. These memos are coded along dimensions such as whether the lease was treated as capital lease or not, and length and penalty clause of lease. So, data is collected on the dependent variable.

---

[3] Facts and values like IFRS, σ, and John Jones are called instances and annotated with single quotes, in lower case, and using underscore for space. When all the arguments for a predicate are instances, then that predicate in of itself is a valid First-Order Logic statement because its truth is provable. So $holds(accounting\_standard(ifrs), \sigma)$ is the formal representation corresponding to the informal question, "In a situational context called 'σ,' is 'ifrs' an accounting standard?". This question is answerable.

However in theorem proving, the value of the dependent variable is inferred from values of independent variables; so data collection is required on independent variables. We make assumptions in order to re-state H1a as an ontology axiom where truth values of predicates that correspond to dependent variables are inferred.

Here then are some predicates of the Auditor Decision-Making Ontology:

$$holds(accounting\_standard(As), S^4). \quad (3)$$
$$holds(accounting\_standard\_type(As, Ast), S). \quad (4)$$
$$holds(auditor(A), S). \quad (5)$$
$$holds(has\_auditor\_orientation(A, Ao), S). \quad (6)$$
$$holds(client\_preferred\_treatment(C, Cpt), S). \quad (7)$$
$$holds(client\_preferred\_treatment(C, Cpt), S). \quad (8)$$
$$occurs(audits(A, C), S). \quad (9)$$

Here are some examples of how these predicates are related to the terms, *deliberate_theory_reference*, *desire*, and *has_evidence*, which are representations from the YODA@ssb BDI Ontology.

$$\forall As \forall S\ holds(accounting\_standard(As, S) \rightarrow \\ holds(deliberate\_theory\_reference(As), S). \quad (10)$$

$$\exists A \forall Ao \forall S\ holds(has\_auditor\_orientation(\\ A, Ao), S) \rightarrow holds(desire(Ao), S). \quad (11)$$

$$\exists C \exists B \forall Cpt\ holds(client\_preferred\_treatment(\\ C, Cpt), S \rightarrow holds(has\_evidence(B,\\ client\_preferred\_treatment, Cpt), S). \quad (12)$$

If As is an accounting standard in a situational context S, then it is a deliberate reference. If there is an auditor A with an orientation Ao in S, then Ao is a desire. Finally, if there exists a client C with a preferred treatment Cpt in S, then there exists a belief B about the client's preferred treatment, where evidence is provided that their preferred treatment's value is Cpt.

Subject to the assumptions, H1b can then be stated as follows:

$$\forall A \forall As \forall S \exists C \exists Sc\ holds(accounting\_standard(\\ As), S) \land holds(accounting\_standard\_type(As\\ , principles\_based), S) \land holds(auditor(A), S)\\ has\_auditor\_orientation(A,$$
$$principles\_oriented), S)\\ \land holds(client\_preferred\_treatment(C,\\ opportunistic), Sc) \land S = occurs(audits(A, C), Sc)\\ \rightarrow holds(enforces\_preferred\_treatment(A\\ , nonopportunistic), S). \quad (13)$$

If a client C would prefer an opportunistic treatment in a context Sc and that context changes to a subsequent context S as a result of an auditor A auditing C, and A is a principles-oriented auditor and As is a principles-based accounting standard used in S, then the auditor will enforce a nonopportunistic treatment in S. This axiom does not directly operationalize H1b, but rather operationalizes Jamal and Tan's insight that when the standard used and the auditor are both principles-based, the client realizes that the auditor has a lot of power to negotiate a nonopportunistic treatment. The client tends to prepare a less opportunistic report even if their initial preference was opportunistic, which is what H1a predicts and is supported by the authors' results. By virtue of these 13 representations, we provide a case study demonstration of how management theories can be operationalized using ontologies.

A predicate attains a truth value either by inference using axioms or its truth value is declared. In our ontology, $holds(has\_auditor\_orientation(john\_jones, principles\_oriented), \sigma)$ is simply declared true and this fact may be applied in axiom (13) to prove $holds(enforces\_preferred\_treatment(\\ john\_jones, conservative), \sigma)$ is true. When a predicate with instances as arguments is declared to be true, that predicate is said to be populated. Referring to step 4 in Figure 1, a set of populated predicates constitutes a populated model. A populated model is devoid of axioms so it can be implemented in, say, a relational database. Along this vein, a situational context can be implemented as a database instance. Then, situational context can be modeled as the totality of populated predicates and other relevant data.

The implication is important for practical ontology-driven analytics. Data from a relational database, not some specialized data source, provide input for inference. Ontologies don't need to be developed in close concert with the database, thus increasing likelihood of adoption.

## 4. Summary and Future Work

In this paper, we posit that operationalization of management theories as ontologies is novel *and* do-able. We then demonstrate a case study development of an Auditor Decision-Making Ontology using a pre-existing, YODA@ssb Belief-Desires-Intentions (BDI) Ontology. We also posit that such operationalization can be useful because automated inference using theories can be used as a data analytics technique. The promise is that management theories expressed in journal articles can be directly applied in an ontology-driven analytics application to support a management decision in practice.

---

[4] In First-Order Logic, arguments in predicates in statements (e.g. axioms or declarative statements) that are quantified with a "for all" (∀) or "there exists" (∃) are variables. Our convention is to show variables as starting with a capital letter, so As and S are variables. Predicates in of themselves as presented here are not statements because without being quantified their truth values cannot be proven.

We have demonstrated the development of an ontology, but have left the demonstration of its utility and verification and validation for future work. Moreover, this concept is very preliminary, and there are many possible impediments to its further development. It may be that most management hypotheses are so abstractly stated that they cannot be concretely applied to real-life data without unrealistic assumptions. Simple First-Order Logic employed in our approach may be limiting given that hypotheses reflect statistical significance, not categorical truths. Having said that, future work using probabilistic reasoning and truth maintenance systems could mitigate this limitation. Even if this concept has merit, it may be that management theorists may find it too technically difficult, and data analysts and analytics managers may still find management theories too "theoretical" and largely irrelevant to practitioners.

Nevertheless, the idea that an information technology, namely ontologies, can potentially actualize the academic cliché, "From Theory to Practice," and be applicable to the burgeoning domain of data analytics is novel and exciting.